 \documentclass[wcp,twocolumn,10pt]{jmlr} 

\usepackage{booktabs}
\usepackage[load-configurations=version-1]{siunitx} 

\theorembodyfont{\upshape}
\theoremheaderfont{\scshape}
\theorempostheader{:}
\theoremsep{\newline}

\jmlryear{2021}
\jmlrworkshop{Machine Learning for Health (ML4H) 2021} 

\title[Rationale production to support clinical decision-making]{Rationale production to support clinical decision-making}
 \author{%
 \Name{Niall Taylor} \Email{niall.taylor@psych.ox.ac.uk}\\
 \addr University of Oxford, United Kingdom
 \AND
 \Name{Lei Sha} \Email{lei.sha@cs.ox.ac.uk}\\
 \addr University of Oxford, United Kingdom
 \AND
 \Name{Dan Joyce} \Email{dan.joyce@psych.ox.ac.uk}\\
 \addr University of Oxford, United Kingdom
 \AND
 \Name{Thomas Lukasiewicz} \Email{thomas.lukasiewicz@cs.ox.ac.uk}\\
 \addr University of Oxford, United Kingdom
 \AND
 \Name{Alejo Nevado-Holgado} \Email{alejo.nevado-holgado@psych.ox.ac.uk}\\
 \addr University of Oxford, United Kingdom
 \AND
 \Name{Andrey Kormilitzin} \Email{andrey.kormilitzin@psych.ox.ac.uk}\\
 \addr University of Oxford, United Kingdom
 }

\begin{document}

\maketitle

\vspace{-0.4cm}

\begin{abstract}
 The development of neural networks for clinical artificial intelligence (AI) is reliant on interpretability, transparency, and performance. The need to delve into the black-box neural network and derive interpretable explanations of model output is paramount. A task of high clinical importance is predicting the likelihood of a patient being readmitted to hospital in the near future to enable efficient triage. With the increasing adoption of electronic health records (EHRs), there is great interest in applications of natural language processing (NLP) to clinical free-text contained within EHRs. In this work, we apply InfoCal \citep{Sha2020}, the current state-of-the-art model that produces extractive rationales for its predictions, to the task of predicting hospital readmission using hospital discharge notes. We compare extractive rationales produced by InfoCal to competitive transformer-based models pretrained on clinical text data and for which the attention mechanism can be used for interpretation. We find each presented model with selected interpretability or feature importance methods yield varying results, with clinical language domain expertise and pretraining critical to performance and subsequent interpretability.   

\end{abstract}
\begin{keywords}
Extractive rationales, NLP, adversarial training, hospital readmission, decision support
\end{keywords}

\vspace{-0.4cm}

\section{Introduction}
\label{sec:intro}

The use of machine learning in clinical settings is becoming widespread, and with the adoption of electronic health records (EHRs) has produced big health data and been opened to modern machine learning approaches \citep{Jiang2017, vaci2020natural}. Whilst ubiquitous, there is still a relative lack of uptake in the use of machine learning in live clinical settings, with neural networks often labelled black boxes. The transparency and explainability of neural networks is critical to trust and acceptance in clinical environments as useful tools. 

Research has sought to provide improved interpretability of NLP models using methods to derive extractive rationales for neural networks predictions by the model itself \citep{bastings-etal-2019-interpretable, lei-etal-2016-rationalizing, Sha2020}. The aim of rationale production is to increase explainability of models by extracting the minimal crucial input required to make a class prediction. In NLP, the rationales are subsets of the input text which maintain predictive power. One of the ways rationales have been achieved is by creating a two-module network, i.e., a selector followed by a predictor, which are trained jointly \citep{lei-etal-2016-rationalizing}. Given an input $x$, the selector picks a subset of the input features $r(x)$ (the rationale) by specifying a distribution over the possible rationales. The predictor acts as a standard classifier, taking as input $r(x)$ and predicting a class $\hat{y}$ to compare with the ground truth class $y$. \citet{Sha2020} proposed InfoCal, an improved type of selector-predictor model that uses an information calibration technique and an additional guider module trained jointly with the selector and predictor in an adversarial manner. InfoCal achieves the current state-of-the-art in rationale extraction on tasks such as sentiment analysis and legal judgment prediction, hence we are interested to see how it performs in the medical domain. 


In this work, we apply InfoCal to the task of predicting hospital readmission from EHRs \citep{Johnson2016}. We compare InfoCal with clinical domain Bidirectional Encoder Transformer (BERT) models, ClinicalBERT \citep{Huang2019} and BioClinicalBERT \citep{Alsentzer2019}. Additionally we compare InfoCal exrractive rationales with importance features in the BERT models via self-attention and layerwise relevance propagation (LRP). We find that the BERT models outperform InfoCal on the classification task, but has a relatively limited mechanism for interpretability in the form of self-attention. InfoCal was able to produce extractive rationales which reach baseline performance on the classification task, and we argue the difficulty lies in the domain expertise created by pretraining present in the BERT based models. 

\vspace{-0.4cm}

\section{Related Work}
\label{sec:EHRs}
With the advent of big data and machine learning, research is beginning to glean insights from the many types of EHRs data \citep{li2020behrt,Huang2019,Kormilitzin2020,Weng2017, Wang2018, Kuruvilla2014, Barak-Corren, Johnson2016}. Data within EHRs can be either structured (following a pre-defined data structure and type, such as ECG recordings, x-ray images, laboratory results, and demographics) or unstructured data (lacking formal rules, type, and bounds, such as the free-text clinical notes which remain in a natural language format). The frequency and volume in which clinical notes are recorded for individual patients surpass that of any other data type within EHRs, with some patients having hundreds of individual notes over their entire history of care \citep{Huang2019,Boag_etal_2020,Barak-Corren,Wang2018, Weng2017}. Consequently, the information within these clinical notes arguably provides a nuanced, rich picture of a patient's symptoms and trajectory to augment other measures. 
Other research has applied similar selector-predictor style models to medical text for certain tasks, such as diagnosis code prediction,and  treatment success inference \citep{mullenbach2018,Lehman2019}. To our knowledge this work is the first to apply InfoCal to a clinical domain task, and to compare rationales with self-attention and layerwise relevance propagation (LRP) in BERT models.  

\vspace{-0.4cm}

\section{Experiments}
\label{sec:methods}
\paragraph{Dataset.}
We use the Medical Information Mart for Intensive Care III (MIMIC-III) \citep{Johnson2016}, a widely-used, freely available dataset developed by the MIT Lab for Computational Physiology. It comprises of de-identified health data associated with 38,597 critical care patients and 58,976 intensive care unit (ICU) admissions at the Beth Israel Deaconess Medical Center between 2001 and 2012. It includes demographics, vital signs, laboratory tests, medications, caregiver notes, imaging reports, and mortality in and out of hospital. 
Our task is to predict the likelihood of an individual patient being re-admitted to the hospital within 30 days of their initial entry to the ICU. To model only unexpected readmission, we exclude any patients who revisited the hospital due to a scheduled appointment. Any admissions with deaths are also removed, and we limit the cohort to non-newborns only, because newborn patients are routinely transferred and discharged frequently, leading to a different distribution of clinical notes and readmission when compared with the general population. The final study cohort contains 34,560 patients with 2,963 positive readmission labels and 42,358
negative readmission labels. 

\begin{figure}[!htb]
    \centering
    \includegraphics[width=0.5\textwidth]{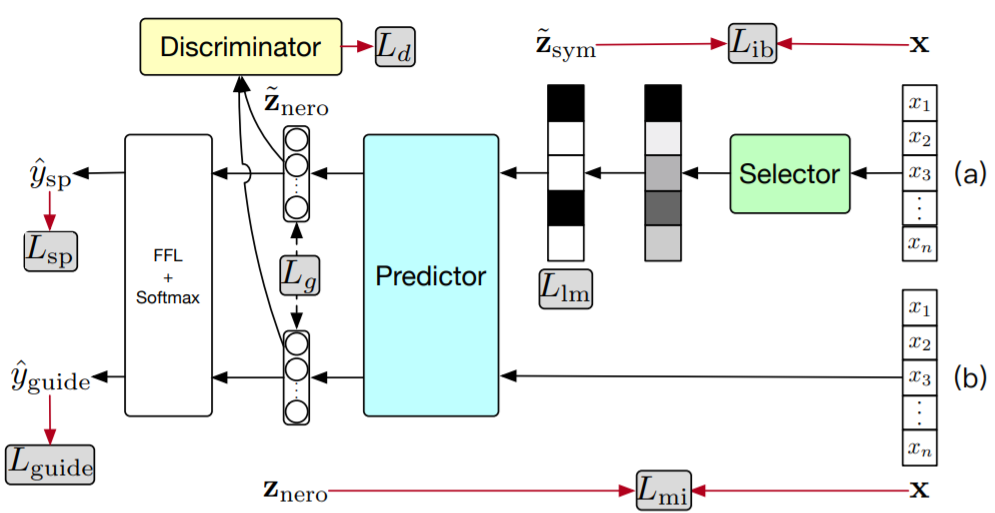}
    \caption{Architecture of InfoCal. The grey boxes are the loss functions, with the red arrows indicating data required to calculate loss. Taken from \citep{Sha2020} }
    \label{fig:InfoCal}
\end{figure}

\paragraph{InfoCal.}
 This is a modular architecture that utilises adversarial training and information bottleneck to produce extractive rationales to justify the models' predictions. A top-level overview is provided, see Fig.\ref{fig:InfoCal}. The key components are the selector, guider, predictor, and discriminator modules. The goal of the selector and guider is to generate a subset of features or tokens to provide to the predictor using the information bottleneck technique. These selected features are further calibrated through training or fooling a discriminator model via adversarial training. The discriminator is attempting to distinguish the generated features in the selector model from the full set given to the guider model. The goal is to encourage the selected features to contain minimal yet sufficient information from the full set. An important note here is that the original InfoCal paper had also used a pretrained language model regularizer to encourage fluency and coherence in extracted rationales, whereby the rationales produced incurred a higher loss if they did not conform with the language models representation of the corpus: however, this reduced performance in our case, thus was ommited.

For a full overview of the model architecture and mathematical foundations, please see \citep{Sha2020}. For training details, see appendix.

\paragraph{BERT models.} We use two pretrained Bidirectional Encoder Representations from Transformer (BERT) models; ClinicalBERT \citep{Huang2019}, pretrained on MIMIC III clinical text only, initialised from the original BERT model \citep{devlin2019bert} and BioClinicalBERT \citep{Alsentzer2019}, pretrained on all MIMIC III notes data, but initialised from BioBERT \citep{LeeBioBert}, for further details see appendix. 

\paragraph{Baseline models.} The following machine learning models were also trained on the classification task to compare with the neural networks performance: logistic regression, random forest, and xgboost. Features to the model were bag of words vector representations of token sequences, for details see appendix. 

\vspace{-0.4cm}

\section{Results}
Evaluation metrics for ClinicalBERT \citep{Huang2019}, BioClinicalBERT \citep{Clark2019}, InfoCal \citep{Sha2020}, and the baseline models are presented in Table \ref{tab:machines}.
A simple visual comparison of interpratibility techniques applied to ClinicalBERT and our InfoCal derived features or rationale is presented in Table \ref{tab:rations}. The techniques used for ClinicalBERT involve looking at self-attention weights of particular layers with respect to the input tokens \citep{devlin2019bert, Huang2019, Clark2019}, and layer-wise relevant propagation, which captures the influence of each input on the model output through back-propagation \citep{Bach2015}, see appendix for further details. 

\label{sec:results}

\begin{table*}[!htb]
  \centering

  \caption{Evaluation metrics of each model. Area under the curve (AUC), area under the precision-recall curve (AUPRC), recall at precision 80 (R80), and F1-Macro scores are presented}
  
  \label{tab:machines}
  \begin{tabular}{rrrrrr}
\toprule
Model      &WordEmbedding &AUC      &AUPRC        &R80      &F1-Macro          \\
\midrule 
clinical-BERT &clinical-BERT   &\textbf{0.805}    &\textbf{0.793}       &\textbf{0.438}      &\textbf{0.702}             \\
BioClinicalBERT &BioClinicalBERT     &0.723    &0.681  &0.124        &0.663                  \\
Logistic Regression &Bag of words &0.702  &0.664   &0.189  &0.641                         \\
Random Forest   &Bag of words    &0.703    &0.686    &0.234    &0.643           \\
XGBoost &Bag of words   &0.702     &0.681       &0.053    &0.633                \\
\midrule
\textbf{Info-Cal}  &word2vec    &0.689     &0.688     &0.152             &0.644               \\

\bottomrule
\end{tabular}
\end{table*}

\begin{table*}[!htb]
\begin{center}
\caption{Comparison of different model derived explanations for an example clinical discharge passage. The selected tokens, highlighted in blue, were those given most weight, or attention by the model in the case of ClinicalBERT, and the selected features or rationale in the case of InfoCal}
\label{tab:rations}
\begin{tabular}{p{0.1\textwidth}p{0.1\textwidth}p{0.7\textwidth}}
\hline
\textbf{Model} & \textbf{Prediction} &\textbf{Sample text} \\
\hline

      clin-BERT with self-attention & Readmission &Has experienced acute on chronic \textcolor[rgb]{0.5543118699137254,0.6900970112156862,0.9955155482352941}{\textbf{diastolic}} heart failure in the setting of volume overload due to his sepsis  \textcolor[rgb]{0.5543118699137254,0.6900970112156862,0.9955155482352941}{\textbf{prescribed warfarin}} due to high sys blood pressure  160 setting\\
      \midrule
     clin-BERT with LRP & Readmission &Has experienced acute on chronic diastolic  \textcolor[rgb]{0.5543118699137254,0.6900970112156862,0.9955155482352941}{\textbf{heart failure}} in the setting of volume \textcolor[rgb]{0.5543118699137254,0.6900970112156862,0.9955155482352941}{\textbf{overload}} due to his sepsis prescribed warfarin due to high sys blood pressure 160 \textcolor[rgb]{0.5543118699137254,0.6900970112156862,0.9955155482352941}{\textbf{setting}} \\
     \midrule
     \textbf{InfoCal rationales} & Readmission &Has experienced acute on chronic diastolic heart failure in the setting of volume overload due to his sepsis prescribed \textcolor[rgb]{0.5543118699137254,0.6900970112156862,0.9955155482352941}{\textbf{warfarin}} due to \textcolor[rgb]{0.5543118699137254,0.6900970112156862,0.9955155482352941}{\textbf{high}} sys blood \textcolor[rgb]{0.5543118699137254,0.6900970112156862,0.9955155482352941}{\textbf{pressure}} 160 setting \\
     \bottomrule

    \end{tabular}
  \end{center}
\end{table*}

\vspace{-0.4cm}

\section{Discussion}

\paragraph{Predicting hospital readmission.}

The novel InfoCal model achieved results in line with much simpler baseline models and the literature \citep{Huang2019, Barbieri2020}. However, the ClinicalBERT model still outperforms all models significantly. Although, it must be stated that the reported performance of InfoCal reflects predictions based solely on the generated rationale as input to the classifier, which was, on average, substantially fewer tokens than the raw input used by other models presented.  We suggest that the relative inability to apply InfoCal to a clinical domain task is due to the lack of medical text pre-training, and also the underlying word vector embeddings and predictor model used.   

\paragraph{Explaining predictions.}

 Each of the presented interpretability methods shown in Table \ref{tab:rations} highlight a number of tokens as most important in the models processing and subsequent output. There is some overlap, but it is interesting to see how different approaches yield different features.  For the ClinicalBERT model, we explored self-attention weights and layer-wise relevance propagation, which both work in a layer-by-layer manner, and are post-hoc. Self-attention visualisations highlight the attention weights between tokens in the input sequence and can provide insight into the language representation of the BERT model, but do not necessarily highlight direct links between input and classification output. Layer-wise Relevance Propagation (LRP) on the other hand acts like a forward pass in reverse, using the weights and neuron activation as a signal of specific input relevance to the final output \citep{Bach2015, arras-etal-2016-explaining}. Both approaches are utilising the entire input sequence, and provide quite intuitive results, although there are doubts to the usefulness of attention weights as explainability metrics \citep{wiegreffe-pinter-2019-attention}. Furthermore, with large models such as BERT, using LRP can become quite challenging, with a need to design appropriate propagation rules for different neural network layers \citep{Montavon2019}. We argue models with built-in explainability methods may provide more straightforward interpretability, however, without medical domain expertise found in ClinicalBERT the InfoCal model struggles to extract the most useful information. 

The extracted features or rationales produced by the InfoCal model was shown to achieve solid performance on predicting readmission, although significantly worse than the BERT based models. The InfoCal model intends to yield the least-but-enough information in the form of rationales, which are then used alone to make a prediction on the task at hand. However, it is clear these rationales lack fluency and are often not comprehensible alone, which has been shown to be the case in other applications \citep{kaur2020interpreting, Alufaisan_Marusich_Bakdash_Zhou_Kantarcioglu_2021}. We must reiterate that the original InfoCal work \citep{Sha2020} had used a pretrained language model as a regularizer, which had not worked in our experiment. 
We feel there is certainly promise in the method presented, with extractive rationales extracted and used by the model. But it is clear the state-of-the-art performance of InfoCal in other domains, did not transfer to the clinical domain and task presented. Further work is needed to extend the presented results, for instance, to improve medical domain performance, components of the InfoCal architecture could be integrated with BERT models. There is a strong possibility that the performance shown by the BERT models is transferable to the InfoCal architecture. Whether or not InfoCal can simply be plugged into a BERT model remains to be seen. To fully investigate the usefulness and interpretability of a models supposed explanations or feature salience will likely require the production of clinician derived ground truths.

\acks{NT is supported by the EPSRC Center for Doctoral Training in Health Data Science (EP/S02428X/1), DWJ, ANH and AK are supported by the National Institute for Health Research (NIHR) Artificial Intelligence for Health and Social Care Award (AI-AWARD02183).}

\bibliography{references}

\appendix

\section{Training details}\label{apd:first}

\subsection{InfoCal - Word embeddings and training details}
The InfoCal model was trained for $100$ epochs with a learning rate initialized at $0.001$ and scheduled to decay by 0.01 every 25 epochs. The predictor model $Pred_{sym}$ and guider model $Pred_{G}$  both used recurrent-neural-networks (RNNs) as encoders outputting the final hidden state, with a single classification layer with sigmoid activation function at the end. Word embeddings for the entire network were pretrained $100$ dimension word2vec embedding vectors using the gensim python library. See Fig.13 in appendix.

\paragraph{Language model regularizer. } As stated in the text body, the original InfoCal paper had pretrained a language model on their domain corpuses to act as a regularization technique. The extracted rationales were presented to the pretrained language model during training, whereby loss would decrease if words conformed with the language model. The idea being that the language models has a general representation of the order of words in that domain, and a rationale that incurs a high loss would imply a reduction in fluency and comprehensiveness. 

\subsection{BERT model training details}

\paragraph{Models.} The presented BERT models utilised in this work were ClinicalBERT \citep{Huang2019} and BioClinicalBERT \citep{Clark2019}. ClinicalBERT is initialised from one of the original BERT models produced called "bert-base-uncased" \citep{devlin2019bert}, and pretrained further using clinical text from MIMIC III. BioClinicalBERT was similarly pretrained using clinical text from MIMIC III, but initialised from BioBERT, which had been trained on a much larger medical text corpus of approximately 17 billion words derived from PubMed abstracts and PMC full text articles \citep{LeeBioBert}
\paragraph{Training.} Once pre-training has taken place, the models weights are saved and can be loaded and adapted to any applicably downstream task. The typical approach for developing a BERT classifier from pretrained weights is to add an additional fully-connected layer on top, with the output dependent on the number of classes or task at hand, as in any standard neural network classifier. With the pretrained clinicalBERT and BioClinicalBERT  models, we add a classification head on top, consisting of 3 layers with the following dimensions: ${768 x 2048, 2048  768, 768 x 1}$. Learning rate was set to $2 x 10^-5$. Typically on fine-tuning tasks, BERT models can overfit very quickly, therefore we train for minimal epochs using validation accuracy and loss for early stopping. 

\subsection{Baseline models}
Using the scikit-learn python package, we produce a bag of words (BoW) feature extractor to produce token sequence vector representations with an N-gram size of 1 (word level). All models are initialised using default hyperparameters produced by the scikit-learn package.
\section{Interpretability methods}\label{apd:second}

\subsection{Self-attention}
\label{sec:attention_func}
The attention function in BERT models uses the encoded embeddings associated with each token in the input sequence. A set of queries, keys, and values are constructed by multiplying each embedding by a learned set of weights. These queries, keys and values are passed to the attention function. The attention function outputs a weighted combination of values for each single query. This function has been shown to capture long-range interactions between any elements in the input sequence up-to 512 tokens. With queries, keys, and values denoted as Q, K, and V respectively, the attention function is as follows: 

\begin{equation} \label{attention_func}
Attention(Q,K,V) = {softmax(\frac{QK^T}{\sqrt{d}}V})
\end{equation}
\subsection{Layerwise Relevance Propagation}
\label{sec:LRP}

There has been a modest boom in explainable machine learning (XML) publications addressing the interpret-ability of  common neural network architectures. One popular approach is layer-wise relevance propagation (LRP) \citep{Bach2015,arras-etal-2016-explaining, Montavon2018}. Briefly, LRP attempts to highlight the influence or relevance of individual input features on the final output of the model. Lets say that a classifier is attempting to map a prediction $f(x)$ to an input $x$, LRP creates a decomposition of $f(x)$ as a sum of differential terms  of the separate input dimensions of $x$. 
This decomposition is produced by a backward pass on the neural network, with each neurons associated relevance (magnitude of contribution) redistributed to preceding neurons. LRP is described as a conservative technique whereby the output $y$ is conserved through the backward pass and equal to the sum of the relevance mapping R of the input.

\begin{equation} \label{LRP}
R_{j} = \underset{k}{\sum}\frac{a_{j}w_{jk}}{\sum_{0,j}a_{j}w_{jk}}R_{k} 
\end{equation}

 The calculation the relevance of each neuron associated with an output back to its input, the following LRP rule is followed iteratively, whereby the influence of neuron j on neuron k is calculated. Where a denotes the activation of the neuron and w the weight applied between the two neurons. (for full details see \citep{Bach2015}).

\section{Rationales}

Figure showing the proportion of full input sequence selected as rationale by InfoCal for model decisions: true positives, false positives, true negatives, and false negatives.
\begin{figure}[!htb]
    \centering
    \includegraphics[width=0.5\textwidth]{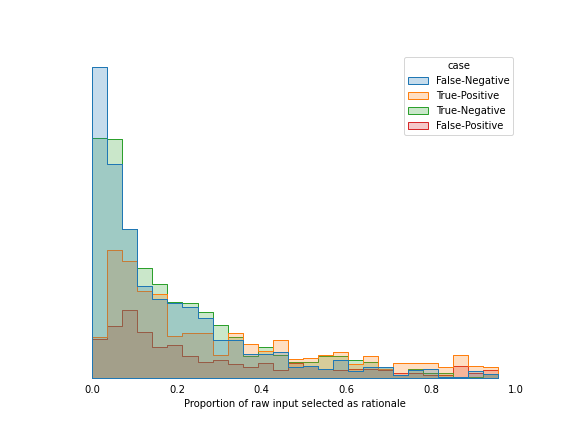}
    \caption{Overlapping histograms of the proportion of chosen words from the original text input that make up the rationale grouped by the classification result: True positive, False positive, True negative, False negative }
    \label{fig:hist_props}
\end{figure}

\end{document}